\documentclass[letterpaper]{article}
\usepackage[utf8]{inputenc}
\usepackage{uai2020}
\usepackage[margin=1in]{geometry}
\usepackage{times}
\usepackage[T1]{fontenc}
\usepackage[ruled]{algorithm2e} 
\usepackage{graphicx}
\usepackage{url}
\usepackage{hyperref}
\usepackage{amsmath}
\usepackage{amsthm}
\usepackage{subcaption}

\theoremstyle{definition}
\newtheorem{definition}{Definition}
\newtheorem{example}{Example}

\usepackage{listings}

\usepackage{todonotes}
\newcounter{todocounter}
\newcommand{\dt}[1]{\stepcounter{todocounter}{\color{purple}  } }
\newcommand{\yz}[1]{\stepcounter{todocounter}{\color{orange}  } }
\newcommand{\hy}[1]{\stepcounter{todocounter}{\color{blue}  } }

\newcommand{\appropto}{\mathrel{\vcenter{
  \offinterlineskip\halign{\hfil$##$\cr
    \propto\cr\noalign{\kern1pt}\sim\cr\noalign{\kern-1pt}}}}}

\lstdefinelanguage{Golang}%
  {morekeywords=[1]{package,import,func,type,struct,return,defer,panic,%
     recover,select,var,const,iota,},%
   morekeywords=[2]{string,uint,uint8,uint16,uint32,uint64,int,int8,int16,%
     int32,int64,bool,float,float32,float,complex64,complex128,byte,rune,uintptr,%
     error,interface},%
   morekeywords=[3]{map,slice,make,new,nil,len,cap,copy,close,true,false,%
     delete,append,real,imag,complex,chan,},%
   morekeywords=[4]{for,break,continue,range,go,goto,switch,case,fallthrough,if,%
     else,default,},%
   morekeywords=[5]{Println,Printf,Error,Print,},%
   sensitive=true,%
   morecomment=[l]{//},%
   morecomment=[s]{p*}{*/},%
   morestring=[b]',%
   morestring=[b]",%
   morestring=[s]{`}{`},%
}

\lstdefinelanguage{Stan}{
  morekeywords=[1]{functions,data,parameters,transformed,model,generated,quantities,%
    for,in,while,print,if,else,lower,upper,increment_log_prob,T,return,%
    reject,integrate_ode,integrate_ode_bdf,integrate_ode_rk45,target},%
  morekeywords=[2]{int,real,vector,%
    ordered,positive_ordered,simplex,unit_vector,%
    row_vector,matrix,%
    cholesky_factor_corr,cholesky_factor_cov,%
    coor_matrix,cov_matrix,%
    void},%
  morekeywords=[3]{%
    Phi,%
    Phi_approx,%
    abs,%
    acos,%
    acosh,%
    append_col,%
    append_row,%
    asin,%
    asinh,%
    atan,%
    atan2,%
    atanh,%
    bernoulli_cdf,%
    bernoulli_cdf_log,%
    bernoulli_lccdf,%
    bernoulli_lcdf,%
    bernoulli_logit_lpmf,%
    bernoulli_logit_lpmf,%
    bernoulli_lpmf,%
    bernoulli_lpmf,%
    bernoulli_rng,%
    bessel_first_kind,%
    bessel_second_kind,%
    beta_binomial_cdf,%
    beta_binomial_cdf_log,%
    beta_binomial_lccdf,%
    beta_binomial_lcdf,%
    beta_binomial_lpmf,%
    beta_binomial_lpmf,%
    beta_binomial_rng,%
    beta_cdf,%
    beta_cdf_log,%
    beta_lccdf,%
    beta_lcdf,%
    beta_lpdf,%
    beta_lpdf,%
    beta_rng,%
    binary_log_loss,%
    binomial_cdf,%
    binomial_cdf_log,%
    binomial_coefficient_log,%
    binomial_lccdf,%
    binomial_lcdf,%
    binomial_logit_lpmf,%
    binomial_logit_lpmf,%
    binomial_lpmf,%
    binomial_lpmf,%
    binomial_rng,%
    block,%
    categorical_logit_lpmf,%
    categorical_logit_lpmf,%
    categorical_lpmf,%
    categorical_lpmf,%
    categorical_rng,%
    cauchy_cdf,%
    cauchy_cdf_log,%
    cauchy_lccdf,%
    cauchy_lcdf,%
    cauchy_lpdf,%
    cauchy_lpdf,%
    cauchy_rng,%
    cbrt,%
    ceil,%
    chi_square_cdf,%
    chi_square_cdf_log,%
    chi_square_lccdf,%
    chi_square_lcdf,%
    chi_square_lpdf,%
    chi_square_lpdf,%
    chi_square_rng,%
    cholesky_decompose,%
    col,%
    cols,%
    columns_dot_product,%
    columns_dot_self,%
    cos,%
    cosh,%
    crossprod,%
    csr_extract_u,%
    csr_extract_v,%
    csr_extract_w,%
    csr_matrix_times_vector,%
    csr_to_dense_matrix,%
    cumulative_sum,%
    determinant,%
    diag_matrix,%
    diag_post_multiply,%
    diag_pre_multiply,%
    diagonal,%
    digamma,%
    dims,%
    dirichlet_lpdf,%
    dirichlet_lpdf,%
    dirichlet_rng,%
    distance,%
    dot_product,%
    dot_self,%
    double_exponential_cdf,%
    double_exponential_cdf_log,%
    double_exponential_lccdf,%
    double_exponential_lcdf,%
    double_exponential_lpdf,%
    double_exponential_lpdf,%
    double_exponential_rng,%
    e,%
    eigenvalues_sym,%
    eigenvectors_sym,%
    erf,%
    erfc,%
    exp,%
    exp2,%
    exp_mod_normal_cdf,%
    exp_mod_normal_cdf_log,%
    exp_mod_normal_lccdf,%
    exp_mod_normal_lcdf,%
    exp_mod_normal_lpdf,%
    exp_mod_normal_lpdf,%
    exp_mod_normal_rng,%
    expm1,%
    exponential_cdf,%
    exponential_cdf_log,%
    exponential_lccdf,%
    exponential_lcdf,%
    exponential_lpdf,%
    exponential_lpdf,%
    exponential_rng,%
    fabs,%
    falling_factorial,%
    fdim,%
    floor,%
    fma,%
    fmax,%
    fmin,%
    fmod,%
    frechet_cdf,%
    frechet_cdf_log,%
    frechet_lccdf,%
    frechet_lcdf,%
    frechet_lpdf,%
    frechet_lpdf,%
    frechet_rng,%
    gamma_cdf,%
    gamma_cdf_log,%
    gamma_lccdf,%
    gamma_lcdf,%
    gamma_lpdf,%
    gamma_lpdf,%
    gamma_p,%
    gamma_q,%
    gamma_rng,%
    gaussian_dlm_obs_lpdf,%
    gaussian_dlm_obs_lpdf,%
    get_lp,%
    gumbel_cdf,%
    gumbel_cdf_log,%
    gumbel_lccdf,%
    gumbel_lcdf,%
    gumbel_lpdf,%
    gumbel_lpdf,%
    gumbel_rng,%
    head,%
    hypergeometric_lpmf,%
    hypergeometric_lpmf,%
    hypergeometric_rng,%
    hypot,%
    if_else,%
    inc_beta,%
    int_step,%
    inv,%
    inv_chi_square_cdf,%
    inv_chi_square_cdf_log,%
    inv_chi_square_lccdf,%
    inv_chi_square_lcdf,%
    inv_chi_square_lpdf,%
    inv_chi_square_lpdf,%
    inv_chi_square_rng,%
    inv_cloglog,%
    inv_gamma_cdf,%
    inv_gamma_cdf_log,%
    inv_gamma_lccdf,%
    inv_gamma_lcdf,%
    inv_gamma_lpdf,%
    inv_gamma_lpdf,%
    inv_gamma_rng,%
    inv_logit,%
    inv_phi,%
    inv_sqrt,%
    inv_square,%
    inv_wishart_lpdf,%
    inv_wishart_lpdf,%
    inv_wishart_rng,%
    inverse,%
    inverse_spd,%
    is_inf,%
    is_nan,%
    lbeta,%
    lchoose,%
    lgamma,%
    lkj_corr_cholesky_lpdf,%
    lkj_corr_cholesky_lpdf,%
    lkj_corr_cholesky_rng,%
    lkj_corr_lpdf,%
    lkj_corr_lpdf,%
    lkj_corr_rng,%
    lmgamma,%
    lmultiply,%
    log,%
    log10,%
    log1m,%
    log1m_exp,%
    log1m_inv_logit,%
    log1p,%
    log1p_exp,%
    log2,%
    log_determinant,%
    log_diff_exp,%
    log_falling_factorial,%
    log_inv_logit,%
    log_mix,%
    log_rising_factorial,%
    log_softmax,%
    log_sum_exp,%
    logistic_cdf,%
    logistic_cdf_log,%
    logistic_lccdf,%
    logistic_lcdf,%
    logistic_lpdf,%
    logistic_lpdf,%
    logistic_rng,%
    logit,%
    lognormal_cdf,%
    lognormal_cdf_log,%
    lognormal_lccdf,%
    lognormal_lcdf,%
    lognormal_lpdf,%
    lognormal_lpdf,%
    lognormal_rng,%
    machine_precision,%
    max,%
    mdivide_left_tri_low,%
    mdivide_right_tri_low,%
    mean,%
    min,%
    modified_bessel_first_kind,%
    modified_bessel_second_kind,%
    multi_gp_cholesky_lpdf,%
    multi_gp_cholesky_lpdf,%
    multi_gp_lpdf,%
    multi_gp_lpdf,%
    multi_normal_cholesky_lpdf,%
    multi_normal_cholesky_lpdf,%
    multi_normal_cholesky_rng,%
    multi_normal_lpdf,%
    multi_normal_lpdf,%
    multi_normal_prec_lpdf,%
    multi_normal_prec_lpdf,%
    multi_normal_rng,%
    multi_student_t_lpdf,%
    multi_student_t_lpdf,%
    multi_student_t_rng,%
    multinomial_lpmf,%
    multinomial_lpmf,%
    multinomial_rng,%
    multiply_log,%
    multiply_lower_tri_self_transpose,%
    neg_binomial_2_cdf,%
    neg_binomial_2_cdf_log,%
    neg_binomial_2_lccdf,%
    neg_binomial_2_lcdf,%
    neg_binomial_2_log_lpmf,%
    neg_binomial_2_log_lpmf,%
    neg_binomial_2_log_rng,%
    neg_binomial_2_lpmf,%
    neg_binomial_2_lpmf,%
    neg_binomial_2_rng,%
    neg_binomial_cdf,%
    neg_binomial_cdf_log,%
    neg_binomial_lccdf,%
    neg_binomial_lcdf,%
    neg_binomial_lpmf,%
    neg_binomial_lpmf,%
    neg_binomial_rng,%
    negative_infinity,%
    normal_cdf,%
    normal_cdf_log,%
    normal_lccdf,%
    normal_lcdf,%
    normal_lpdf,%
    normal_lpdf,%
    normal_rng,%
    not_a_number,%
    num_elements,%
    ordered_logistic_lpmf,%
    ordered_logistic_lpmf,%
    ordered_logistic_rng,%
    owens_t,%
    pareto_cdf,%
    pareto_cdf_log,%
    pareto_lccdf,%
    pareto_lcdf,%
    pareto_lpdf,%
    pareto_lpdf,%
    pareto_rng,%
    pareto_type_2_cdf,%
    pareto_type_2_cdf_log,%
    pareto_type_2_lccdf,%
    pareto_type_2_lcdf,%
    pareto_type_2_lpdf,%
    pareto_type_2_lpdf,%
    pareto_type_2_rng,%
    pi,%
    poisson_cdf,%
    poisson_cdf_log,%
    poisson_lccdf,%
    poisson_lcdf,%
    poisson_log_lpmf,%
    poisson_log_lpmf,%
    poisson_log_rng,%
    poisson_lpmf,%
    poisson_lpmf,%
    poisson_rng,%
    positive_infinity,%
    pow,%
    prod,%
    qr_Q,%
    qr_R,%
    quad_form,%
    quad_form_diag,%
    quad_form_sym,%
    rank,%
    rayleigh_cdf,%
    rayleigh_cdf_log,%
    rayleigh_lccdf,%
    rayleigh_lcdf,%
    rayleigh_lpdf,%
    rayleigh_lpdf,%
    rayleigh_rng,%
    rep_array,%
    rep_matrix,%
    rep_row_vector,%
    rep_vector,%
    rising_factorial,%
    round,%
    row,%
    rows,%
    rows_dot_product,%
    rows_dot_self,%
    scaled_inv_chi_square_cdf,%
    scaled_inv_chi_square_cdf_log,%
    scaled_inv_chi_square_lccdf,%
    scaled_inv_chi_square_lcdf,%
    scaled_inv_chi_square_lpdf,%
    scaled_inv_chi_square_lpdf,%
    scaled_inv_chi_square_rng,%
    sd,%
    segment,%
    sin,%
    singular_values,%
    sinh,%
    size,%
    skew_normal_cdf,%
    skew_normal_cdf_log,%
    skew_normal_lccdf,%
    skew_normal_lcdf,%
    skew_normal_lpdf,%
    skew_normal_lpdf,%
    skew_normal_rng,%
    softmax,%
    sort_asc,%
    sort_desc,%
    sort_indices_asc,%
    sort_indices_desc,%
    sqrt,%
    sqrt2,%
    square,%
    squared_distance,%
    step,%
    student_t_cdf,%
    student_t_cdf_log,%
    student_t_lccdf,%
    student_t_lcdf,%
    student_t_lpdf,%
    student_t_lpdf,%
    student_t_rng,%
    sub_col,%
    sub_row,%
    sum,%
    tail,%
    tan,%
    tanh,%
    tcrossprod,%
    tgamma,%
    to_array_1d,%
    to_array_2d,%
    to_matrix,%
    to_row_vector,%
    to_vector,%
    trace,%
    trace_gen_quad_form,%
    trace_quad_form,%
    trigamma,%
    trunc,%
    uniform_cdf,%
    uniform_cdf_log,%
    uniform_lccdf,%
    uniform_lcdf,%
    uniform_lpdf,%
    uniform_lpdf,%
    uniform_rng,%
    variance,%
    von_mises_lpdf,%
    von_mises_lpdf,%
    von_mises_rng,%
    weibull_cdf,%
    weibull_cdf_log,%
    weibull_lccdf,%
    weibull_lcdf,%
    weibull_lpdf,%
    weibull_lpdf,%
    weibull_rng,%
    wiener_lpdf,%
    wiener_lpdf,%
    wishart_lpdf,%
    wishart_lpdf,%
    wishart_rng
  },%
  otherkeywords={<-,~,+=,=},%
  sensitive=true,%
  morecomment=[l]{\#},%
  morecomment=[l]{//},%
  morecomment=[n]{/*}{*/},%
  string=[d]"
  literate={<-}{{$\leftarrow$}}1 {~}{{$\sim$}}1%
}

\title{Stochastically Differentiable Probabilistic Programs}
\author{David Tolpin \\
Ben-Gurion University\\
of the Negev
\And
Yuan Zhou \\
University of Oxford
\And
Hongseok Yang \\
Korea Advanced Institute\\
of Science and Technology
}

\begin{document}

\maketitle

\begin{abstract}
	Probabilistic programs with mixed support (both continuous and discrete 
	latent random variables) commonly appear in many probabilistic programming systems (PPSs). 
	However, the existence of the discrete random variables prohibits many basic gradient-based inference engines, which makes the inference procedure on such models particularly challenging. 
	Existing PPSs either require the user to manually marginalize out the discrete variables
	or to perform a composing inference by running inference
	separately on discrete and continuous variables.
	The former is infeasible in most cases whereas the latter 
	has some fundamental shortcomings. 
	We present a novel approach to run inference efficiently and robustly
	in such programs using stochastic gradient Markov Chain Monte Carlo family of
	algorithms.
	We compare our stochastic gradient-based inference algorithm against conventional baselines
	in several important cases of probabilistic programs with mixed support,
    and demonstrate that it outperforms existing composing inference baselines
    and works almost as well as inference in marginalized
    versions of the programs, 
    but with less programming effort and
    at a lower computation cost.
\end{abstract}

\section{Introduction}

Probabilistic programming \cite{GMR+08,MSP14,WVM14,GS15}
represents statistical models as programs written in an
otherwise general programming language that provides syntax for
the definition and conditioning of random variables.  Inference
can be performed on probabilistic programs to obtain the
posterior distribution or point estimates of the variables.
Inference algorithms are provided by the probabilistic
programming framework, and each algorithm is usually applicable
to a wide class of probabilistic programs in a black-box automated manner.
The algorithms include Markov Chain Monte Carlo (MCMC) variants
--- Metropolis-Hastings~\cite{WSG11,MSP14,YHG14}, Hamiltonian
Monte Carlo (HMC)~\cite{N12,Stan17}, as well as expectation
propagation~\cite{MWG+10}, extensions of Sequential Monte
Carlo~\cite{WVM14,MYM+15,PWD+14,RNL+2016,MS18}, variational
inference~\cite{WW13,KTR+17}, and gradient-based
optimization~\cite{Stan17,BCJ+19}.

A probabilistic program computes the (unnormalized) probability of
its execution~\cite{WVM14}. An execution is summarized as an
instantiation of the program \textit{trace}, and the probability
is a function of the trace.  Some probabilistic programming
frameworks require that the trace shape be specified
upfront, i.e. static~\cite{Stan17,T19}, while others allow introduction of trace
components dynamically~\cite{GMR+08,P09,GS15,GXG18,TMY+16}, in
the course of a program execution.

Efficient inference methods for probabilistic programs,
especially in the high-dimensional scenario,
often involve computation of the gradient with respect to the
latent variables~\cite{S16,Stan17,GXG18,T19,BCJ+19}. 
This restricts gradient-based inference methods to differentiable programs
with continuous variables only.  
Probabilistic programs with a
mixture of continuous and discrete variables, such as mixture
models, state space models, and factor models, must
resort to alternatives which do not rely solely on
differentiability, at the cost of lower performance and poorer
scalability.  

Take a standard Gaussian mixture model (GMM) as an example.
In GMM, the continuous latent variables are usually the mean and
variance of the Gaussian distribution of each mixture component,
and the discrete variables are the assignments of each data point to a
mixture component. 
Inference in GMM can be performed by either manually
marginalizing out the discrete latent variables, or by combining
conditional gradient-based inference on continuous variables,
and conditional gradient-free (such as Gibbs sampling) inference
on discrete variables. 
Manual marginalization is straightforward in the case of GMM, as suggested by Stan~\cite{S16}, 
but can be more complex in most other cases, and even intractable.  
Conditional inference on
each group of variables is prone to slow mixing and poor
robustness in the face of multimodality.
A generic robust
inference method which both exploits differentiability and is
applicable to probabilistic programs with mixed support is
highly desirable.

In this work, we focus on the probabilistic programs with a
mixture of continuous and discrete variables, differentiable
with respect to the continuous variables,
and propose to treat them as \textit{stochastically differentiable}. 
We derive an \emph{unbiased}
stochastic estimate of the gradient of the marginal log likelihood that can be efficiently computed.
We then demonstrate how one can apply stochastic gradient-based
inference methods~\cite{MCF15}, in particular stochastic
gradient Hamiltonian Monte Carlo (sgHMC)~\cite{CFG14}, 
on stochastically differentiable models with this estimate.
To show the potential usage,
a reference implementation of the probabilistic programming facility
that supports stochastically differentiable probabilistic
programs is presented in Section~\ref{sec:studies}.
We compare our proposed adaptation of sgHMC on three stochastically differentiable models
against manually marginalized inference 
and the state-of-the-art automated method in PPSs, composing inference, 
and empirically confirm the substantial improvements of our approach.

\paragraph{Contributions} This work brings the following contributions:
\begin{itemize}
	\vspace{-5pt}
    \item the notion of a stochastically differentiable
        probabilistic program;
    \vspace{-3pt}
    \item an \emph{unbiased} stochastic estimator for the gradient of the marginal log likelihood of such a program;
    \vspace{-3pt}
    \item an adaptation of sgHMC with this estimator as the automated inference engine;
    \vspace{-3pt}
    \item a reference implementation of probabilistic
        programming with support for stochastically
        differentiable probabilistic programs in Go programming language~\cite{Golang}.
\end{itemize}

\paragraph{Notation} We denote by $p(\pmb{x})$ the probability
or probability density of random variable $\pmb{x}$, and by
$p(\pmb{x}|\pmb{y})$ the conditional probability or probability
density of $\pmb{x}$ given $\pmb{y}$, depending on the domain of
$\pmb{x}$.  We write $\pmb{x} \sim p(\cdot)$ when $\pmb{x}$ is a
random variable with probability $p(\cdot)$. We denote by
$\tilde p(\cdot)$ a probability known up to a normalization
constant, i.e. the \emph{unnormalized} probability.

\section{Stochastically Differentiable Probabilistic Program}
\label{sec:spp}

To define a stochastically differentiable probabilistic program,
we begin with definitions of a \textbf{deterministic} and a
\textbf{stochastic}
probabilistic program. Different definitions of (deterministic)
probabilistic programs are given in the literature and reflect
different views and accents. For the purpose of this work, let
us give the following broad definition:
\begin{definition}A deterministic probabilistic program $\Pi$
    that defines a distribution over traces $p(\pmb{x}|\pmb{y})$ is a
    computer program that accepts a \textit{trace assignment}
    $\pmb{x}$ as one of its arguments, and returns the
    unnormalized probability of $\pmb{x}$ conditioned on other
    arguments $\pmb{y}$:
    \begin{equation}
        \Pi(\pmb{x}, \pmb{y}) \Rightarrow \tilde p(\pmb{x}|\pmb{y}).
    \end{equation}
\end{definition}

The trace assignment $\pmb{x}$ may have the form of a vector, or of
a list of address-value pairs, or any other form suitable
for a particular implementation. Accordingly, let us define a
stochastic probabilistic program:
\begin{definition} A stochastic probabilistic program $\Xi$
    that defines a distribution over traces $p(\pmb{x}|\pmb{y})$ is a
    computer program that accepts a \textit{trace assignment}
    $\pmb{x}$ as one of its arguments, and returns the
    unnormalized probability of $\pmb{x}$ conditioned on other
    arguments $\pmb{y}$ and random variable $\pmb{z}$
    conditioned on $\pmb{y}$:
    \begin{align}
        \label{eqn:spp}
        \pmb{z} & \sim p(\pmb{z}|\pmb{y}) \\ \nonumber
        \Xi(\pmb{x}, \pmb{y}) & \Rightarrow \tilde p(\pmb{x}|\pmb{y},\pmb{z}).
    \end{align}
    \label{def:spp}
\end{definition}

\hy{One thing that confused me many times is that the distribution of $\pmb{z}$
can depend on $\pmb{y}$. Since the condition distribution of $x$ is not normalised,
the distribution of $\pmb{z}$ is not quite a posterior, but it is also not
a prior either because of its dependency on the observed $\pmb{y}$. Maybe
explaining the rationale behind this design decision may help some readers.}

A rationale for this definition is that $\pmb{z}$ corresponds to
the \emph{nuisance} parameters or nondeterministic choices inside the
program. Finally, let us define a stochastically differentiable
probabilistic program:
\begin{definition} A stochastically differentiable probabilistic
    program $\Xi$ is a stochastic probabilistic program
    (Definition~\ref{def:spp}) with trace $\pmb{x} \in
    \mathcal{R}^n$ such that $\tilde p(\pmb{x}|\pmb{y},\pmb{z})$
        is differentiable w.r.t. $\pmb{x}$ for any
        $\pmb{y}$ and $\pmb{z}$.
    \label{def:sdpp}
\end{definition}


\begin{figure*}[!t]
	\centering
	\hspace{5pt}
		\begin{subfigure}{0.45\textwidth}
			\centering
    \begin{lstlisting}[basicstyle=\small]
 func Survey(theta float, y []bool) float {
     prob := Beta(1, 1).pdf(theta)
     for i := 0; i != len(y); i++ {
         coin := Bernoulli(0.5).sample()
         if coin {
             prob *= Bernoulli(theta).cdf(y[i])
         } else {
             prob *= Bernoulli(0.5).cdf(y[i])
         }
      }
      return prob
  }\end{lstlisting}
  \subcaption{A probabilistic program for the compensation
  	survey. Random variables \lstinline{coin} at line $4$ are the nuisance variables
  	 which make the program stochastic.}
  \label{fig:survey-stochastic}
		\end{subfigure}%
	~
	\hspace{15pt}
	\begin{subfigure}{0.45\textwidth}
		\centering
	\vspace{-10pt}
    \begin{lstlisting}[basicstyle=\small]
 func Survey(theta float, y []bool) float {
     prob := Beta(1, 1).pdf(theta)
     for i := 0; i != len(y); i++ {
         prob *= 0.5*Bernoulli(theta).cdf(y[i])
                +0.5*Bernoulli(0.5).cdf(y[i])
     }
     return prob
 }\end{lstlisting}
 \subcaption{An equivalent deterministic version of the compensation
 	survey model in Figure~\ref{fig:survey-stochastic}.
 	The probability is marginalized over the coin flip
 	in the code of the program.}
 \label{fig:survey-deterministic} 
	\end{subfigure}%
\caption{Compensation survey model}
\end{figure*}

Let us illustrate a stochastic (and stochastically
differentiable) probabilistic program on an example: 
\begin{example}
    A survey is conducted among a company's employees. The survey
    contains a single question: ``Are you satisfied with your
    compensation?'' To preserve employees' privacy, the employee
    flips a coin before answering the question. If the coin
    shows head, the employee answers honestly; otherwise, the
    employee flips a coin again, and answers `yes' on head, `no'
    on tail. Based on survey outcomes, we want to know how many
    of the employees are satisfied with their compensations.
    \label{ex:survey}
\end{example}

\paragraph{Stochastic probabilistic program} The stochastic
probabilistic program modelling the survey
(Figure~\ref{fig:survey-stochastic}) receives two parameters:
probability \lstinline{theta} that a randomly chosen employee is
satisfied with the compensation and vector \lstinline{y} of
boolean survey outcomes. The program trace consists of a single
variable \lstinline{theta}.  There are nuisance random variables
\lstinline{coin} representing coin flips which are sampled
inside the program from their prior distribution, but are not
included in the trace --- this makes the probabilistic program
stochastic. The unnormalized probability of the trace is
accumulated in variable \lstinline{prob}.  First, a Beta prior
is imposed on \lstinline{theta} (line 2).  Then,
\lstinline{prob} is multiplied by the probability of each answer
given \lstinline{theta} and \lstinline{coin} (lines 5--9).

\paragraph{Manual marginalization} Though not generally the
case, the program in
Figure~\ref{fig:survey-stochastic} can be rewritten as a
deterministic probabilistic program
(Figure~\ref{fig:survey-deterministic}). Instead of flipping a
coin as in line~4 of the stochastic program in
Figure~\ref{fig:survey-stochastic}, the probability of
observation \lstinline{y[i]} is computed in lines~5--6 as the
sum of probabilities given either head or tail, weighted by the
probabilities of head and tail (both are 0.5).
For case studies (Section~\ref{sec:studies}), we chose
probabilistic programs for which manual marginalization was
possible, and compared inference in stochastic and 
marginalized versions.

\yz{probably add a section on background: HMC; MH+HMC; sgHMC}

\section{Inference}
\label{sec:infer}

Efficient posterior inference in probabilistic models for which
a stochastic gradient estimate of the logarithm of the
unnormalized posterior density is available can be performed
using a stochastic gradient Markov Chain Monte
Carlo method~\cite{MCF15}. Stochastic gradient posterior inference was
originally motivated by the handling of large data sets, where the
gradient was estimated by subsampling --- computing the
gradient on a small part of the dataset~\cite{CFG14}, but
stochastic gradient MCMC methods are agnostic to the way the
gradient is estimated, given an unbiased estimate is available,
and can as well be applied to stochastically differentiable
probabilistic programs. However, a naive estimate of the
gradient as $\nabla_{\pmb{x}} \log \tilde p(\pmb{x}|\pmb{y}, \pmb{z})$,
where $\pmb{z}$ is drawn from $p(\pmb{z}|\pmb{y})$, 
is a \textbf{biased} 
estimate of the gradient of $\nabla_{\pmb{x}} \log
\tilde p(\pmb{x}|\pmb{y})$. 
\yz{not exactly sure why this is biased and why the following is not}
\dt{because the above is a stochastic gradient estimate of the
product integral rather than of the regular (sum) integral; it
corresponds to the case when we express non-determinism/know the
posterior distribution, rather than marginalize on the prior
distribution; the original paper on stochastic probabilistic
programs, on arxiv, discusses both cases and tries to outline
the difference between them. Just technically, you get an
unbiased estimate if there is a sum over a set of samples, and
you compute (and scale) the gradient of a sub-sum. For
marginalization, there is $\log$ sum of samples, so you cannot
just take the gradient of one term (or a few  terms) under the
logarithm, and use it as an unbiased estimate.}
\yz{Thanks, David. I think i probably did not phrase the question well. 
What I was confused was actually what is biased. 
To be very explicit, 
the following estimate of the gradient is biased, ie
$$
\nabla_{\pmb{x}} \log \tilde p(\pmb{x}|\pmb{y})  \approx \frac 1 N \sum_{i=1}^N \nabla_{\pmb{x}} \log \tilde p(\pmb{x}|\pmb{y}, \pmb{z}_i) \\
$$
where $\nonumber \pmb{z}_i \sim p(\pmb{z}|\pmb{y})$ which is the prior of $z$. 
The point is we cannot take the same calculation as in $p(x, y)$ by marginalizing $z$ out from the prior. 
btw, why does the prior of $z$, which is $p(\pmb{z}|\pmb{y})$, depend on $y$?

In fact, I found this terminology a bit confusing. 
I would like to suggest to use $p(\pmb x, \pmb y)$ instead of $\tilde p(\pmb{x}|\pmb{y})$ for the unnormalized density.
Especially when we have all x, y and z. Eg $\tilde p(\pmb{x}|\pmb{y}, \pmb{z})$ actually means $p(\pmb{x}, \pmb{y}| \pmb{z})$.
}
\hy{The bias comes from the use of logarithm. The RHS of your approximate
equation but without the gradient is a biased estimator of the LHS without
the gradient.}

In what follows, we derive an
\textbf{unbiased} estimate of the stochastic gradient, and elaborate on
the use of the estimate within the framework of stochastic
gradient Hamiltonian Monte Carlo~(sgHMC).

\subsection{Unbiased Stochastic Gradient Estimate}

The unnormalized probability density $\tilde p(\pmb{x}|\pmb{y})$
is a marginalization of $\tilde p(\pmb{x}|\pmb{y},\pmb{z})$ over
$\pmb{z}$:
\begin{equation}
    \tilde p(\pmb{x}|\pmb{y}) = \int_z p(\pmb{z}|\pmb{y}) \tilde p(\pmb{x}|\pmb{y},\pmb{z})d\pmb{z}
\end{equation}
Posterior inference and maximum \textit{a posteriori} estimation
require a stochastic estimate of $\nabla_{\pmb{x}} \log \tilde p(\pmb{x}|\pmb{y})$:
\begin{align}
    \nonumber \nabla_{\pmb{x}} \log \tilde p(\pmb{x}|\pmb{y}) &= \frac {\int_z p(\pmb{z}|\pmb{y}) \tilde p(\pmb{x}|\pmb{y},\pmb{z})\nabla_{\pmb{x}} \log \tilde p(\pmb{x}|\pmb{y},\pmb{z})d\pmb{z}} {\tilde p(\pmb{x}|\pmb{y})}  \\
    \nonumber &= \int_z p(\pmb{z}|\pmb{y}) \frac {\tilde p(\pmb{x}|\pmb{y},\pmb{z})} {\tilde p(\pmb{x}|\pmb{y})} \nabla_{\pmb{x}} \log \tilde p(\pmb{x}|\pmb{y},\pmb{z})d\pmb{z} \\ \nonumber
    &= \int_z  \frac {p(\pmb{x}|\pmb{y},\pmb{z})p(\pmb{z}|\pmb{y})} {p(\pmb{x}|\pmb{y})} \nabla_{\pmb{x}} \log \tilde p(\pmb{x}|\pmb{y},\pmb{z})d\pmb{z} \\
    &= \int_z p(\pmb{z}|\pmb{x},\pmb{y}) \nabla_{\pmb{x}} \log \tilde p(\pmb{x}|\pmb{y},\pmb{z})d\pmb{z}
\end{align}
By Monte Carlo approximation,
\begin{align}
    \label{eqn:grad-mc-marg}
    \nonumber \pmb{z}_i & \sim p(\pmb{z}|\pmb{x}, \pmb{y}) \\
    \nabla_{\pmb{x}} \log \tilde p(\pmb{x}|\pmb{y}) & \approx \frac 1 N \sum_{i=1}^N \nabla_{\pmb{x}} \log \tilde p(\pmb{x}|\pmb{y}, \pmb{z}_i)
\end{align}
Draws of $\pmb{z}_i$ are conditioned on $\pmb{x}$ and can be
approximated by a Monte Carlo method, such as Markov chain Monte
Carlo; 
$p(\pmb{z}|\pmb{y})$ need not be known for the
approximation. 
Note that there are two Monte Carlo
approximations involved: 
\begin{enumerate}
    \item gradient $\nabla_{\pmb{x}} \log \tilde p(\pmb{x}|\pmb{y})$
        is approximated by a finite sum of gradients
        $\nabla_{\pmb{x}} \log \tilde p(\pmb{x}|\pmb{y}, \pmb{z}_i)$;
    \item draws of $\pmb{z}_i$ from $p(\pmb{z}|\pmb{x}, \pmb{y})$ are
        approximated.
\end{enumerate}

An intuition behind estimate (\ref{eqn:grad-mc-marg}) is that
for marginalization, assignments to nuisance parameters which
make the trace assignment more likely contribute more to the
gradient estimate.

\subsection{Inference with Stochastic Gradient HMC}

Stochastic gradient Hamiltonian Monte Carlo~\cite{CFG14} does
not involve Metropolis-Hastings correction and requires only
that a routine that computes a stochastic estimate of the
gradient of the unnormalized posterior probability density should be
provided. We propose here to use a single-sample gradient
estimate, where $\pmb{z}$ is drawn using an MCMC method, such as
a Gibbs sampler or a variant of Metropolis-Hastings:
\begin{algorithm}[h]
    \caption{Single-sample gradient estimate}
    \label{alg:sge}
    \begin{enumerate}
        \item Draw $\pmb{z}$ from $p(\pmb{z}|\pmb{x}, \pmb{y})$ (MCMC).
        \item Return $\nabla_{\pmb{x}} \log \tilde p(\pmb{x}|\pmb{y}, \pmb{z})$.
    \end{enumerate}
\end{algorithm}

Note that $\pmb{z}$ is freshly drawn for each estimation of the gradient.
Hence, every step through $\pmb{x}$ is performed using a gradient estimate
based on \textit{a new sample} of $\pmb{z}$.
\yz{I am a little bit confused what our algorithm is doing exactly. 
	If I understand sgHMC correctly, it uses a mini-batch stochastic gradient estimator
	in the original paper right. 
	Are we doing exactly the sgHMC?
It feels like we are doing the following: for each $1$ to $L$ Leapfrog step in the sgHMC, we do Alg. 1. 
But we are using Eq.~\ref{eqn:grad-mc-marg} as the gradient estimator on \emph{full} data, i.e. all $y$s, rather than a mini-batch.}
\dt{sgHMC uses an unbiased stochastic gradient estimate to
sample from the posterior but does not restrict how this
estimate is obtained. In the case of 'deep learning'/ large data
sets, the log density is with respect to \textbf{all} samples in
the datasets. So, to compute the log density over all samples
you take the sum of log densities of each sample. The gradient
of log density is the sum of gradients of log densities of each
sample. You estimate the gradient stochastically by selecting a sub
sum uniformly and scaling up.

In our case, we have the same sgHMC algorithm which uses an
unbiased stochastic gradient estimate, but the stochasticity
comes from a different source. We use all of the data
but choose certain assignments to discrete variables for each
gradient computation. In a sense, our data which we subsample is
the set of all samples from the posterior distribution of the
nuisance variables. }
\yz{I see. We should probably point this out as it is not super straightforward if people don't know sgHMC well.
In fact, I will point this out when I write the background section.}

Variance of this stochastic gradient estimate can be reduced by
computing and averaging multiple estimates, each for a new draw
of $\pmb{z}$. This is similar to the well-known 
mini-batch stochastic gradient descent in stochastic optimization. In the case
studies (Section~\ref{sec:studies}), we compare statistical
efficiency and computation time of inference with the
single-sample and multi-sample stochastic gradient estimates.

Compare inference with sgHMC, using Algorithm~\ref{alg:sge} as the
gradient estimate, to alternating inference, with (non-stochastic)
HMC on $\pmb{x}$ and an MCMC variant on $\pmb{z}$:
\begin{algorithm}[h]
    \caption{Baseline: alternating HMC on $\pmb{x}$ and MCMC on $\pmb{z}$}
    \label{alg:hmc}
    \begin{enumerate}
        \item Draw $\pmb{z}$ from $p(\pmb{z}|\pmb{x}, \pmb{y})$ (MCMC).
        \item Draw $\pmb{x}$ from $p(\pmb{x}|\pmb{y}, \pmb{z})$ (HMC).
    \end{enumerate}
\end{algorithm}

Despite apparent similarity between Algorithms~\ref{alg:sge}
and~\ref{alg:hmc}, one HMC iteration of Algorithm~\ref{alg:hmc}
involves multiple computations of the gradient for each draw of
$\pmb{x}$, all for \textit{the same sample} of $\pmb{z}$, 
as in the Leapfrog steps. 
A new sample of
$\pmb{z}$ is only drawn between iterations of HMC. This
results in poorer mixing and may lead the sampler to get stuck
in one of the modes of a multimodal posterior distribution.

\begin{figure*}[!t]
	\centering
	\hspace{5pt}
	\begin{subfigure}{0.45\textwidth}
\begin{lstlisting}
func TwoNormals(x float) float {
    z := Bernoulli(0.5).sample()
    if z {
        return Normal(1, 0.5).pdf(x)
    } else {
        return Normal(-1, 0.5).pdf(x)
    }
}\end{lstlisting}
\subcaption{Mixture of two Gaussians: nuisance variable $z$
	selects either $\mathcal{N}(-1, 0.5)$ or 5$\mathcal{N}(1,
	0.5)$.}
\label{fig:sge-vs-hmc-pp}
	\end{subfigure}%
	~
	\hspace{5pt}
	\begin{subfigure}{0.45\textwidth}
 \includegraphics[width=\linewidth]{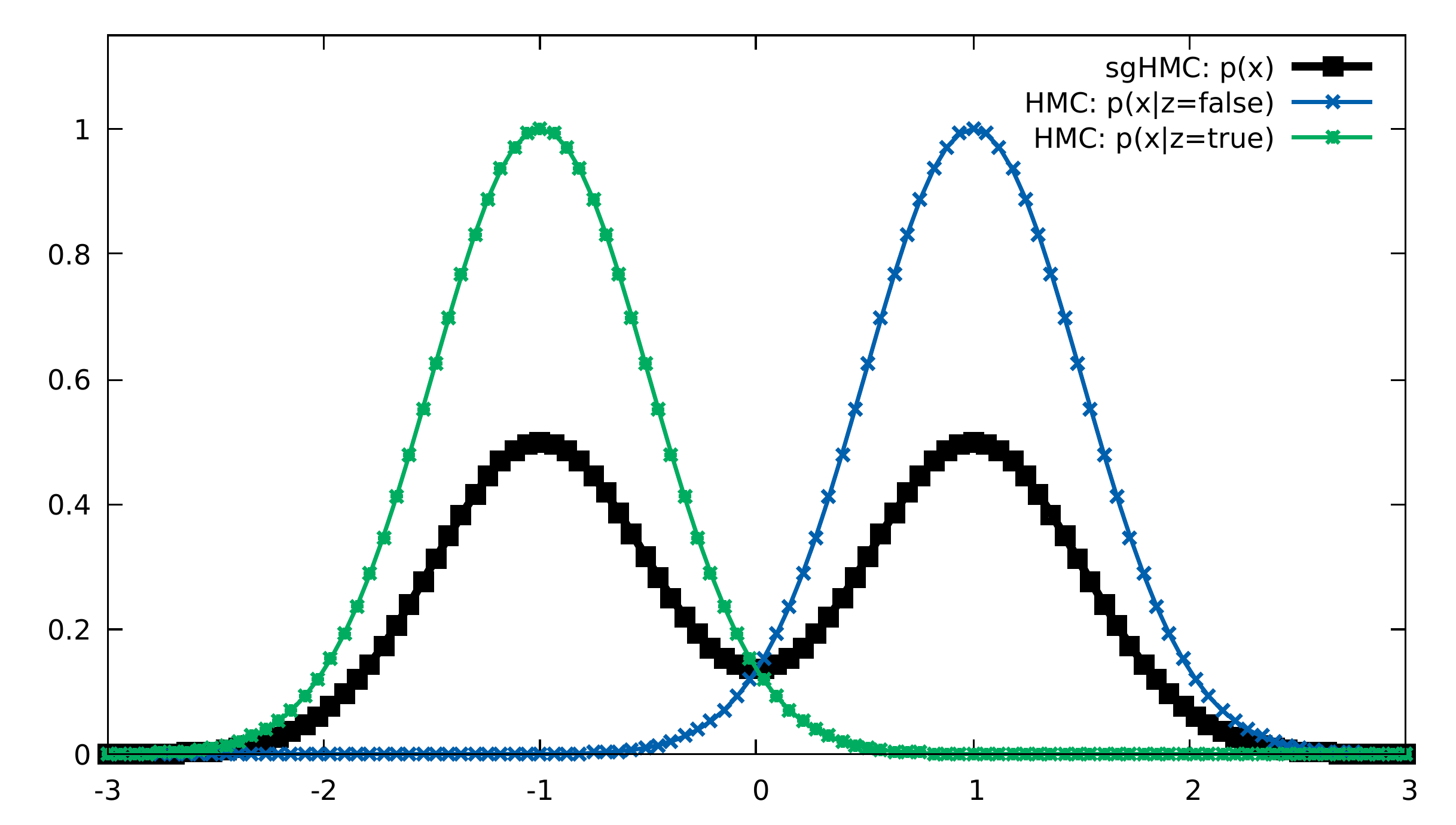}
                \subcaption{Mixture of two Gaussians: HMC conditioned on either $\pmb{z}=\mathit{false}$
                or $\pmb{z}=\mathit{true}$ will alternate between exploring each of the
	components, while sgHMC will directly draw samples from the posterior.}
\label{fig:sge-vs-hmc}
	\end{subfigure}%
	\caption{An illustration example: sgHMC vs. MCMC + HMC}
\end{figure*}

As an illustration, consider the probabilistic program in
Figure~\ref{fig:sge-vs-hmc-pp}.
This program specifies a Gaussian mixture model, $p(x) = 0.5 p(x|\mu=-1, \sigma=0.5) + 0.5
p(x|\mu=1, \sigma=0.5)$. Nuisance variable $z$ selects either
component with equal probability.  Figure~\ref{fig:sge-vs-hmc}
shows densities of each of the components and of the mixture.
For Algorithm~\ref{alg:sge}, \yz{Re-phrase  or make a alg. block: Alg.1 is not a full algorithm}
it directly samples from the entire mixture density, i.e. it re-draws $\pmb z$ before each gradient step of $\pmb x$, which enables the posterior samples of $\pmb x$ to cover both modes.
However, the HMC step of Algorithm~\ref{alg:hmc} will conditioned on a fixed $\pmb z$, 
i.e. sample $\pmb x$ for $L$ leapfrog steps conditioned on $\pmb z = 1$ or $2$, 
which makes the samples mainly concentrated around one mode. 
The estimates of $\pmb x$ will only evolve to the other mode once $\pmb z$ changes
based on samples from the current mode, with potentially low mixing rate.

More evidence for poorer mixing and lower statistical efficiency
is provided in the case studies. Compared to that,
Algorithm~\ref{alg:sge} estimates the gradient for a new draw of
$\pmb{z}$ on each invocation. 

\section{Case Studies}
\label{sec:studies}

We implemented inference in stochastically differentiable
probabilistic programs using Infergo~\cite{T19}, a probabilistic
programming facility for the Go programming language. Due to
reliance of Infergo on pure Go code for probabilistic programs,
no changes were necessary to the way probabilistic programs are
represented. Implementation of sgHMC for the case studies in
this paper will be made publicly available upon publication
and proposed for inclusion into Infergo.

We evaluated inference in stochastically differentiable
probabilistic programs on several models. In each evaluation, we
created two probabilistic programs: 
the standard probabilistic program with a mixed support 
as users may write (akin to Figure~\ref{fig:survey-stochastic}) 
and a corresponding marginalized program (akin to Figure~\ref{fig:survey-deterministic}).
Additionally, for comparison,
we implemented a marginalized program for each
model in Stan~\cite{Stan17}; 
we verified that the posterior
distributions inferred by each of inference schemes on both
mixed support and marginalized programs are consistent with the
Stan version. We provide results for three models in this study:
compensation survey (Section~\ref{sec:survey}),  Gaussian
mixture model, and hidden Markov model. Details of each model
are provided later in this section.  The full source code, data,
and evaluation scripts for the case studies will be made
publicly available upon publication.

The purpose of these case studies is to compare statistical
efficiency and computation time of sgHMC, combination of MCMC
and HMC on mixed-support models, and HMC on marginalized models. 
For each model, we measured, and averaged over 10 independent
runs, the effective sample size and the computation time. Four
inference schemes were applied:
\begin{itemize}
	\item (ours) sgHMC with a single-sample gradient estimate;
	\item (ours) sgHMC with 10-sample gradient estimate;
	\item a combination of MCMC (sitewise Metropolis-Hastings)
		on discrete variables and HMC on continuous variables;
	\item HMC on the marginalized model.
\end{itemize}
All inference schemes were run to produce $10\,000$ samples
(sufficient for convergence on all models in the studies),
with $10$ steps (gradient estimates) between samples. 
The step resulting in the largest effective sample size for HMC on the
marginalized model was chosen and fixed for all over schemes
on the same model. This choice of the step size favors HMC in
comparison, meaning that the actual comparative performance of
sgHMC on stochastically differentiable probabilistic programs is
at least as good, or better, as what follows from the evaluation.

\begin{table*}[!t]
    \centering
    \caption{Effective sample size}
    \label{tab:ess}

    \begin{tabular}{r | c | c | c | c }
        {\it model} & {\it sgHMC, 1 sample} & {\it sgHMC, 10 samples} & {\it MH+HMC} & {\it HMC, marginalized}  \\ \hline
        Survey & $4600\pm 180$ & $5000\pm 160$ & $2800\pm 300$ & $5200\pm 200$  \\
        GMM & $5900\pm240$ & $6800\pm200$ & $1900\pm200$ & $7200\pm150$  \\
        HMM & $6200\pm280$ & $6300\pm220$ & $4800\pm190$ & $6700\pm180$
    \end{tabular}
\end{table*}

\begin{table*}[!t]
    \centering
    \caption{Computation time, seconds}
    \label{tab:time}

    \begin{tabular}{r | c | c | c | c }
        {\it model} & {\it sgHMC, 1 sample} & {\it sgHMC, 10 samples} & {\it MH+HMC} & {\it HMC, marginalized}  \\ \hline
        Survey & $6.5 \pm 0.1$ & $40 \pm 1$ & $7.4\pm0.1$ & $21\pm0.5$ \\
        GMM & $35\pm 0.7$ & $340\pm 5$ & $38\pm 0.7$ &  $95\pm2$ \\
        HMM & $10\pm 0.3$ & $98 \pm 3$ & $10 \pm 0.3$ & $46 \pm 0.8$ 
    \end{tabular}
\end{table*}

\begin{table*}[!t]
	\centering
	\caption{Effective sample size per second}
	\label{tab:persec}
	
	\begin{tabular}{r | c | c | c}
		{\it model} & {\it sgHMC} & {\it MH+HMC} & {\it HMC}  \\ \hline
		Survey & {\bf 650} & 380  & 240 \\
		GMM & {\bf 170} & 50  & 75 \\
		HMM & {\bf 620} & 480 & 145 \\
	\end{tabular}
\end{table*}

The measurements are summarized in Tables~\ref{tab:ess}
and~\ref{tab:time}, including empirical standard deviations over
multiple runs. Table~\ref{tab:ess} shows effective sample sizes.
Larger numbers are better.  Effective sample size of HMC on
marginalized program is 15\%--20\% larger than sgHMC with a
single-sample gradient estimate and 5--7\% larger than sgHMC
with 10-sample estimate. The MH+HMC combination consistently
produces sample sizes which are lower by 30--60\%.
Table~\ref{tab:time} shows computation times. While 10-sample
gradient estimate is computationally expensive and results in
almost linear increase in the total computation time,
cancelling the improvement due to a less noisy gradient
estimate, sgHMC exhibits the shortest computation time for all
models, 2.5--4.5 times shorter than HMC on the marginalized
program. This is because the marginalized version has both a
higher asymptotic complexity (more details are in descriptions
of each of the case studies) and involves computationally
expensive exponentiation (calls to \lstinline{LogSumExp} in
Figures~\ref{fig:survey-marg}, \ref{fig:gmm-marg}, and~\ref{fig:hmm-marg}).

Table~\ref{tab:persec} combines Tables~\ref{tab:ess}
and~\ref{tab:time} to show approximate effective sample size in
seconds. Despite a slightly lower total effective sample size of
sgHMC, the effective sample size of sgHMC with a single-sample
gradient estimate is 2--4 times larger than that of HMC on the
marginalized model. To summarize, sgHMC with a single-sample
gradient estimate has the best performance on the mixed support
probabilistic programs used in the case studies, outperforming
HMC on marginalized models by a number of times even for
parameters favoring HMC in the comparison.

\subsection{Compensation Survey}
\label{sec:survey}


\begin{figure*}[!t]
	\begin{subfigure}{0.5\textwidth}
		\begin{lstlisting}
func (m *StochasticModel) 
	Observe(x []float64) float64 {
  ll := 0.
  theta := mathx.Sigm(x[0])
  for i := 0; i != len(m.Y); i++ {
    if m.Coin[i] {
      ll += Flip.Logp(theta, m.Y[i])
    } else {
      ll += Flip.Logp(0.5, m.Y[i])
    }
  }
  return ll
}\end{lstlisting}
	\subcaption{Compensation survey}
	\label{fig:survey}
	\end{subfigure}%
~
\vspace{5pt}
\begin{subfigure}{0.5\textwidth}
	\begin{lstlisting}
func (m *DeterministicModel) 
	Observe(x []float64) float64 {
  ll := 0.
  theta := mathx.Sigm(x[0])
  for i := 0; i != len(m.Y); i++ {
    ll += mathx.LogSumExp(
      Flip.Logp(theta, m.Y[i])+math.Log(0.5),
      Flip.Logp(0.5, m.Y[i])+math.Log(0.5))
  }
  return ll
}\end{lstlisting}
	\subcaption{Compensation survey, marginalized}
	\label{fig:survey-marg}
\end{subfigure}%
\caption{Code Example for the Compensation Survey}
\end{figure*}

%

The compensation survey follows the description in
Section~\ref{sec:survey}. The mixed support program is shown in
Figure~\ref{fig:survey} and the marginalized version in
Figure~\ref{fig:survey-marg}. For evaluation, a data set of 60
observations, corresponding to 67\% satisfaction, was used.  The
marginalized version has the same asymptotic complexity  as the
mixed support version, but should take at least twice as long to
run due to computation of the log probability for both coin flip
outcomes (lines~5--7).  This is consistent with the empirical
measurements, 21 vs 6.5 seconds, (Table~\ref{tab:time}), with
extra time spent in \lstinline{LogSumExp}.

\subsection{Gaussian Mixture Model}

\begin{figure*}[!t]
	\begin{subfigure}{0.5\textwidth}
	\begin{lstlisting}
func (m *StochasticModel) 
	Observe(x []float64) float64 {
  ll := Normal.Logps(0, 10, x...)
  for j := range m.Mu {
    m.Mu[j] = x[2*j]
    m.Sigma[j] = math.Exp(x[2*j+1])
  }
  for i := 0; i != len(m.Data); i++ {
    j := m.Assgn[i]
    ll += Normal.Logp(m.Mu[j], m.Sigma[j], 
                               m.Data[i])
  }
  return ll
}\end{lstlisting}
	\caption{Gaussian mixture model}
	\label{fig:gmm}
\end{subfigure}%
~
\vspace{5pt}
\begin{subfigure}{0.5\textwidth}
	\begin{lstlisting}
func (m *DeterministicModel) 
	Observe(x []float64) float64 {
  ll := Normal.Logps(0, 10, x...)
  for j := range m.Mu {
    m.Mu[j] = x[2*j]
    m.Sigma[j] = math.Exp(x[2*j+1])
  }
  for i := 0; i != len(m.Data); i++ {
    var l float64
    for j := 0; j != m.NComp; j++ {
      lj := Normal.Logp(m.Mu[j], m.Sigma[j], m.Data[i])
      if j == 0 {
        l = lj
      } else {
        l = mathx.LogSumExp(l, lj)
      }
    }
    ll += l
  }
  return ll
}\end{lstlisting}
	\caption{Gaussian mixture model, marginalized}
	\label{fig:gmm-marg}
	\end{subfigure}%
\caption{Code Example for the GMM}
\end{figure*}

%

The Gaussian mixture model infers means and standard deviation
of the components, given a set of samples from the mixture and
the number of components. The mixed support program is shown in
Figure~\ref{fig:gmm} and the marginalized version in
Figure~\ref{fig:gmm-marg}. For evaluation, a data set of
100 observations, corresponding to two components with equal
probability of either component, was used. The computation time 
of the marginalized version is at least the computation time of
the mixed support program times the number of components (at least
twice longer for two components) due to the inner loop in
lines~8--15. This is consistent with the empirical measurements,
95 vs. 35 seconds (Table~\ref{tab:time}), with extra time spent
in \lstinline{LogSumExp}.

\subsection{Hidden Markov Model}


\begin{figure*}[!t]
	\begin{subfigure}{0.5\textwidth}
	\begin{lstlisting}
func (m *StochasticModel) 
  Observe(x []float64) float64 {
  ll := Normal.Logps(0, 10, x...)
  for i := range m.logT {
    D.SoftMax(x[i*m.NStat:(i+1)*m.NStat], 
	      m.logT[i])
    for j := range m.logT[i] {
      m.logT[i][j] = math.Log(m.logT[i][j])
    }
  }
  noise := math.Exp(m.Noise)
  for i := range m.Data {
    ll+= Normal.Logp(float64(m.States[i]), noise,
                     m.Data[i])
    if i != 0 {
      ll += m.logT[m.States[i-1]][m.States[i]]
    }
  }
  return ll
}\end{lstlisting}
	\subcaption{Hidden Markov model}
	\label{fig:hmm}
\end{subfigure}%
~
\vspace{5pt}
\begin{subfigure}{0.4\textwidth}
	\begin{lstlisting}
func (m *DeterministicModel) 
  Observe(x []float64) float64 {
  ll := Normal.Logps(0, 10, x...)
  for i := range m.logT {
    D.SoftMax(x[i*m.NStat:(i+1)*m.NStat], m.logT[i])
    for j := range m.logT[i] {
      m.logT[i][j] = math.Log(m.logT[i][j])
    }
  }
  noise := math.Exp(m.Noise)
  for i := range m.Data {
    for j := 0; j != m.NStat; j++ {
      m.gamma[i][j] = Normal.Logp(float64(j), noise,
                                  m.Data[i])
      if i != 0 {
        for j_ := 0; j_ != m.NStat; j_++ {
          m.acc[j_] = m.gamma[i-1][j_] + m.logT[j_][j]
        }
        m.gamma[i][j] += D.LogSumExp(m.acc)
      }
    }
  }
  ll += D.LogSumExp(m.gamma[len(m.gamma)-1])
  return ll
}
	\end{lstlisting}
	\subcaption{Hidden Markov model, marginalized}
	\label{fig:hmm-marg}
	\end{subfigure}%
	\caption{Code Example for the HMM}
\end{figure*}

%
%

The Hidden Markov model infers the transition matrix given the
emission matrix and the observations.  The mixed support program
is shown in Figure~\ref{fig:hmm} and the marginalized version in
Figure~\ref{fig:hmm-marg}. For evaluation, a data set of 16
observations, corresponding to three distinct hidden states, was
used. Following~\cite{SDT18}, the marginalized version
implements the costly forward pass of the forward-backward
algorithm. The computation time of the marginalized version, in
a na\"ive implementation, is
at least the computation time of the mixed support program times
the squared number of hidden states (at least nine times longer
for three hidden states). This is due to the doubly-nested loop
in lines~11--20. The forward pass also involves two distinct
calls to \lstinline{LogSumExp}, one in line~18, on each
iteration of the inner loop, and the other in line~22. The
marginalized version in Figure~\ref{fig:hmm-marg} optimizes the
inner loop by reusing computations, but the computation time
of the marginalized version is still five times longer than that of
the mixed support version, 46 vs. 10 seconds (Table~\ref{tab:time}).

\section{Related Work}
\label{sec:related}

Stochastically differentiable probabilistic programs as introduced in Section~\ref{sec:spp}, 
though may be named differently, 
commonly appeared in many different probabilistic programming systems~(PPSs).
However, the mixture of continuous and discrete random variables substantially complicates the inference procedures in PPS as automated engines.
Especially, the inference procedure can become extremely challenging 
if the program trace is not static, i.e. the support of the program is dynamic. 

To perform inference in such models in PPSs such as 
Stan~\cite{Stan17},
which are usually designed around one or two efficient gradient-based inference methods such as HMC,  
the user would need to manually marginalize out the discrete nuisance variables since they violate the prerequisite of the inference engines.
Unfortunately, this is not feasible in most cases and also adds unnecessary burden for the user, which violates the principle of automation in PPS at the first place.

Alternative choices may be PPSs such as PyMC3~\cite{S16}, Turing.jl~\cite{GXG18}, and Gen~\cite{Gen19} 
where one can customize different kernels for different variables as composing inference. 
For example, one can use the Metropolis-within-Gibbs sampler for the nuisance variables and HMC for the remaining continuous ones. 
It is probably the state-of-the-art method for this type of models, especially in the PPSs oriented for the gradient-based inference engines. 
However, as we have discussed in Section~\ref{sec:infer} and empirically confirmed in Section~\ref{sec:studies}, 
this method has some fundamental shortcomings and our approach provides substantial practical improvements.

There are also some other approaches for improving inference performance for this type of models. 
For example, LF-PPL~\cite{zhou2019lf} proposed a novel low-level language for PPS to incorporate more sophisticated inference techniques for piecewise differentiable models;
Stochaskell~\cite{roberts2019reversible} provided a way to perform reversible jump MCMC in PPS; 
\cite{ZYT+19} introduced a new sampling scheme, called Divide, Conquer and Combine, to perform inference in the general purpose PPS by dividing the program into sub-programs and conquering individual one before combing into an overall estimate.
However, LF-PPL imposes some restrictions to the language of the PPS that are not required in our setup, 
whereas the latter two do not exploit gradient information at all.
Therefore, they are less relevant to our approach and we leave the possible connection for future work.

\section{Discussion}
\label{sec:disc}

In this paper, we introduced the notion of a stochastically
differentiable probabilistic program as well as an inference
scheme for such programs. 
We provided a reference implementation
of probabilistic programming with stochastically differentiable
probabilistic programs. 
Stochastically differentiable probabilistic programs facilitate natural specification of
probabilistic models and support efficient inference in the
models, providing a viable alternative to explicit model
marginalization, whether manual or algorithmic.

Our understanding of stochastically differentiable probabilistic
programs, classes of models for which they are best suited, and
inference schemes is still at an early stage and evolving. 
In particular, we are working, towards future publications, on
maximum \textit{a posteriori} estimation in stochastically
differentiable probabilistic programs.  Another research
direction would be selection and adaption of stochastic
gradient-based inference algorithms to probabilistic programming
inference.  Last but not least, we are looking into introducing
support for stochastically differentiable probabilistic programs
into existing probabilistic programming frameworks.


\bibliography{refs}

\begin{thebibliography}{10}

\bibitem{BCJ+19}
Eli Bingham, Jonathan~P. Chen, Martin Jankowiak, Fritz Obermeyer, Neeraj
  Pradhan, Theofanis Karaletsos, Rohit Singh, Paul Szerlip, Paul Horsfall, and
  Noah~D. Goodman.
\newblock Pyro: deep universal probabilistic programming.
\newblock {\em Journal of Machine Learning Research}, 20(28):1--6, 2019.

\bibitem{Stan17}
Bob Carpenter, Andrew Gelman, Matthew Hoffman, Daniel Lee, Ben Goodrich,
  Michael Betancourt, Marcus Brubaker, Jiqiang Guo, Peter Li, and Allen
  Riddell.
\newblock {S}tan: a probabilistic programming language.
\newblock {\em Journal of Statistical Software, Articles}, 76(1):1--32, 2017.

\bibitem{CFG14}
Tianqi Chen, Emily~B. Fox, and Carlos Guestrin.
\newblock Stochastic gradient {H}amiltonian {M}onte {C}arlo.
\newblock In {\em Proceedings of the 31st International Conference on
  International Conference on Machine Learning}, ICML'14, pages
  II--1683--II--1691. JMLR.org, 2014.

\bibitem{Gen19}
Marco~F Cusumano-Towner, Feras~A Saad, Alexander~K Lew, and Vikash~K
  Mansinghka.
\newblock Gen: a general-purpose probabilistic programming system with
  programmable inference.
\newblock In {\em Proceedings of the 40th ACM SIGPLAN Conference on Programming
  Language Design and Implementation}, pages 221--236, 2019.

\bibitem{GXG18}
Hong Ge, Kai Xu, and Zoubin Ghahramani.
\newblock Turing: composable inference for probabilistic programming.
\newblock In {\em International Conference on Artificial Intelligence and
  Statistics, {AISTATS} 2018, 9-11 April 2018, Playa Blanca, Lanzarote, Canary
  Islands, Spain}, pages 1682--1690, 2018.

\bibitem{Golang}
The {G}o team.
\newblock The {G}o programming language.
\newblock \url{http://golang.org/}, 2009.

\bibitem{GS15}
N.~D. Goodman and A.~Stuhlm\"uller.
\newblock {\em The Design and Implementation of Probabilistic Programming
  Languages.}
\newblock 2014.
\newblock electronic; retrieved 2019/3/29.

\bibitem{GMR+08}
Noah~D. Goodman, Vikash~K. Mansinghka, Daniel~M. Roy, Keith Bonawitz, and
  Joshua~B. Tenenbaum.
\newblock {C}hurch: a language for generative models.
\newblock In {\em Proc. of Uncertainty in Artificial Intelligence}, 2008.

\bibitem{KTR+17}
Alp Kucukelbir, Dustin Tran, Rajesh Ranganath, Andrew Gelman, and David~M.
  Blei.
\newblock Automatic differentiation variational inference.
\newblock {\em J. Mach. Learn. Res.}, 18(1):430--474, January 2017.

\bibitem{MCF15}
Yi-An Ma, Tianqi Chen, and Emily Fox.
\newblock A complete recipe for stochastic gradient {MCMC}.
\newblock In C.~Cortes, N.~D. Lawrence, D.~D. Lee, M.~Sugiyama, and R.~Garnett,
  editors, {\em Advances in Neural Information Processing Systems 28}, pages
  2917--2925. Curran Associates, Inc., 2015.

\bibitem{MSP14}
Vikash~K. Mansinghka, Daniel Selsam, and Yura~N. Perov.
\newblock {V}enture: a higher-order probabilistic programming platform with
  programmable inference, 2014.

\bibitem{MWG+10}
T~Minka, J~Winn, J~Guiver, and D~Knowles.
\newblock Infer .net 2.4, {M}icrosoft research {C}ambridge, 2010.

\bibitem{MS18}
Lawrence~M. Murray and Thomas~B. Schön.
\newblock Automated learning with a probabilistic programming language:
  {B}irch.
\newblock {\em Annual Reviews in Control}, 46:29 -- 43, 2018.

\bibitem{N12}
Radford~M. Neal.
\newblock {MCMC} using {H}amiltonian dynamics.
\newblock Published as Chapter 5 of the Handbook of Markov Chain Monte Carlo,
  2011, 2012.

\bibitem{PWD+14}
B.~Paige, F.~Wood, A.~Doucet, and Y.W. Teh.
\newblock Asynchronous anytime sequential {M}onte {C}arlo.
\newblock In {\em Advances in Neural Information Processing Systems}, 2014.

\bibitem{P09}
Avi Pfeffer.
\newblock Figaro: An object-oriented probabilistic programming language.
\newblock In {\em Charles River Analytics Technical Report (2009)}, 2009.

\bibitem{RNL+2016}
Tom Rainforth, Christian~A Naesseth, Fredrik Lindsten, Brooks Paige, Jan-Willem
  van~de Meent, Arnaud Doucet, and Frank Wood.
\newblock Interacting particle {M}arkov chain {M}onte {C}arlo.
\newblock In {\em Proceedings of the 33rd International Conference on Machine
  Learning}, volume~48 of {\em JMLR: W\&CP}, 2016.

\bibitem{roberts2019reversible}
David~A Roberts, Marcus Gallagher, and Thomas Taimre.
\newblock Reversible jump probabilistic programming.
\newblock In {\em The 22nd International Conference on Artificial Intelligence
  and Statistics}, pages 634--643, 2019.

\bibitem{S16}
John Salvatier, Thomas~V. Wiecki, and Christopher Fonnesbeck.
\newblock Probabilistic programming in {P}ython using {PyMC}3.
\newblock {\em {PeerJ} Computer Science}, 2:e55, apr 2016.

\bibitem{SDT18}
{Stan Development Team}.
\newblock {\em Stan Modeling Language User's Guide and Reference Manual,
  Version 2.18.0}.
\newblock 2018.

\bibitem{T19}
David Tolpin.
\newblock Deployable probabilistic programming.
\newblock In {\em Proceedings of the 2019 ACM SIGPLAN International Symposium
  on New Ideas, New Paradigms, and Reflections on Programming and Software},
  Onward! 2019, pages 1--16, New York, NY, USA, 2019. ACM.

\bibitem{TMY+16}
David Tolpin, Jan-{W}illem van~de {M}eent, Hongseok Yang, and Frank Wood.
\newblock Design and implementation of probabilistic programming language
  {A}nglican.
\newblock In {\em Proceedings of the 28th Symposium on the Implementation and
  Application of Functional Programming Languages}, IFL 2016, pages 6:1--6:12,
  New York, NY, USA, 2016. ACM.

\bibitem{MYM+15}
Jan-Willem van~de Meent, Hongseok Yang, Vikash Mansinghka, and Frank Wood.
\newblock Particle {G}ibbs with ancestor sampling for probabilistic programs.
\newblock In {\em Artificial Intelligence and Statistics}, 2015.

\bibitem{WSG11}
David Wingate, Andreas Stuhlm\"uller, and Noah~D. Goodman.
\newblock Lightweight implementations of probabilistic programming languages
  via transformational compilation.
\newblock In {\em Proceedings of the 14th Artificial Intelligence and
  Statistics}, 2011.

\bibitem{WW13}
David Wingate and Theophane Weber.
\newblock Automated variational inference in probabilistic programming, 2013.

\bibitem{WVM14}
Frank Wood, Jan-Willem van~de {M}eent, and Vikash Mansinghka.
\newblock A new approach to probabilistic programming inference.
\newblock In {\em Artificial Intelligence and Statistics}, 2014.

\bibitem{YHG14}
Lingfeng Yang, Pat Hanrahan, and Noah~D Goodman.
\newblock Generating efficient {MCMC} kernels from probabilistic programs.
\newblock In {\em Proceedings of the Seventeenth International Conference on
  Artificial Intelligence and Statistics}, pages 1068--1076, 2014.

\bibitem{zhou2019lf}
Yuan Zhou, Bradley~J Gram-Hansen, Tobias Kohn, Tom Rainforth, Hongseok Yang,
  and Frank Wood.
\newblock {LF-PPL: A Low-Level First Order Probabilistic Programming Language
  for Non-Differentiable Models}.
\newblock In {\em The 22nd International Conference on Artificial Intelligence
  and Statistics}, pages 148--157, 2019.

\bibitem{ZYT+19}
Yuan Zhou, Hongseok Yang, Yee~Whye Teh, and Tom Rainforth.
\newblock Divide, conquer, and combine: a new inference strategy for
  probabilistic programs with stochastic support, 2019.

\end{thebibliography}
\bibliographystyle{plain}

\clearpage
\onecolumn
\appendix


\clearpage
\section{Stan versions of the programs in the case studies}

\subsection{Compensation Survey}

\begin{figure*}[!h]
\begin{lstlisting}[language=Stan]
data {
    int<lower=0> N;
    int<lower=0,upper=1> y[N];
}
parameters {
    real<lower=0,upper=1> theta;
}
model {
    for (n in 1:N) {
        target += log_mix(0.5, 
            bernoulli_lpmf(y[n]|theta),
            bernoulli_lpmf(y[n]|0.5));
    }
}
\end{lstlisting}
\end{figure*}

\subsection{Gaussian Mixture Model}

\begin{figure*}[!h]
\begin{lstlisting}[language=Stan]
data {
    int<lower = 0> N;
    vector[N] y;
}
parameters {
    vector[2] mu;
    real<lower=0> sigma[2];
}
model {
    mu ~ normal(0, 10);
    sigma ~ lognormal(0, 10);
    for (n in 1:N)
        target += log_mix(0.5,
            normal_lpdf(y[n] | mu[1], sigma[1]),
            normal_lpdf(y[n] | mu[2], sigma[2]));
    }
}
\end{lstlisting}
\end{figure*}

\subsection{Hidden Markov model}

\begin{figure*}[!h]
\begin{lstlisting}[language=Stan]
data {
    real<lower=0> noise;
    int<lower=1> K; // number of states
    int<lower=1> N;
    real y[N];
}
parameters {
    simplex[K] theta[K]; 
}
model {
real acc[K];
real gamma[N,K];
for (k in 1:K)
    gamma[1,k] = normal_lpdf(y[0]|k-1, noise);
    for (t in 2:N) {
        for (k in 1:K) {
            for (j in 1:K)
                acc[j] = gamma[t-1,j] + log(theta[j,k])
                    + normal_lpdf(y[t]|k-1, noise);
            gamma[t,k] = log_sum_exp(acc);
        }
    }
    target += log_sum_exp(gamma[N]);
}
\end{lstlisting}
\end{figure*}

\end{document}